\documentclass[10pt,twocolumn,letterpaper]{article}

\usepackage[pagenumbers]{cvpr} 

\newcommand{\bh}{\mathbf{h}}\newcommand{\bH}{\mathbf{H}}

\newcommand{\bK}{\mathbf{K}}

\newcommand{\bM}{\mathbf{M}}
\newcommand{\bn}{\mathbf{n}}\newcommand{\bN}{\mathbf{N}}

\newcommand{\bp}{\mathbf{p}}\newcommand{\bP}{\mathbf{P}}

\newcommand{\bR}{\mathbf{R}}
\newcommand{\bS}{\mathbf{S}}
\newcommand{\bt}{\mathbf{t}}
\newcommand{\bu}{\mathbf{u}}

\newcommand{\bW}{\mathbf{W}}
\newcommand{\bx}{\mathbf{x}}

\newcommand{\cam}[0]{\boldsymbol{T}}
\newcommand{\camCW}[0]{\cam_{CW}}

\newcommand{\rotCW}[0]{\boldsymbol{R}_{CW}}

\newcommand{\tauC}[0]{\boldsymbol{\tau}}

\newcommand{\Exp}[0]{\text{Exp}}
\newcommand{\Log}[0]{\text{Log}}

\newcommand{\gaussians}[0]{\mathcal{G}}

\newcommand{\pd}[2]{\frac{\partial {#1} }{\partial {#2} }}
\newcommand{\mpd}[2]{\frac{\mathcal{D} {#1}}{\mathcal{D} {#2}}}
\newcommand{\se}[1]{\mathfrak{se}(#1)}
\newcommand{\SE}[1]{\boldsymbol{SE}(#1)}

\newcommand{\SO}[1]{\boldsymbol{SO}(#1)}
\newcommand{\identity}[0]{\boldsymbol{I}}

\newcommand{\gtimage}[0]{\bar{I}}
\newcommand{\gtdepth}[0]{\bar{D}}

\newcommand{\thetaC}[0]{\boldsymbol{\theta}}
\newcommand{\rhoC}[0]{\boldsymbol{\rho}}

\definecolor{cvprblue}{rgb}{0.21,0.49,0.74}
\usepackage[pagebackref,breaklinks,colorlinks,allcolors=cvprblue]{hyperref}
\usepackage{booktabs}
\usepackage{multirow}
\usepackage{array}
\usepackage{graphicx}
\usepackage{caption}
\usepackage{booktabs}

\newcommand{\webpage}{{\url{https://muskie82.github.io/4dtam/}}}
\newcommand{\webvideo}{{\url{https://youtu.be/MRGhggLmTF0?si=51bqfAe9pYQNWgf-/}}}

\def\paperID{6705} 
\def\confName{CVPR}
\def\confYear{2025}

\title{4DTAM: Non-Rigid Tracking and Mapping via Dynamic Surface Gaussians}

\author{
Hidenobu Matsuki \qquad  Gwangbin Bae \qquad Andrew J. Davison \vspace{0.5em}\\
Dyson Robotics Laboratory, Imperial College London \\
{\tt\small \{h.matsuki20, g.bae, a.davison\}@imperial.ac.uk} \\ \\
\textbf{Website}: \mbox{{\webpage}} \\
\textbf{Video}: \mbox{{\webvideo}}
}

\begin{document}

\setlength\textfloatsep{2pt}
\setlength\intextsep{2pt}
\setlength\abovecaptionskip{2pt}

\twocolumn[{%
\renewcommand\twocolumn[1][]{#1}%
\maketitle
\includegraphics[width=1.0\textwidth]{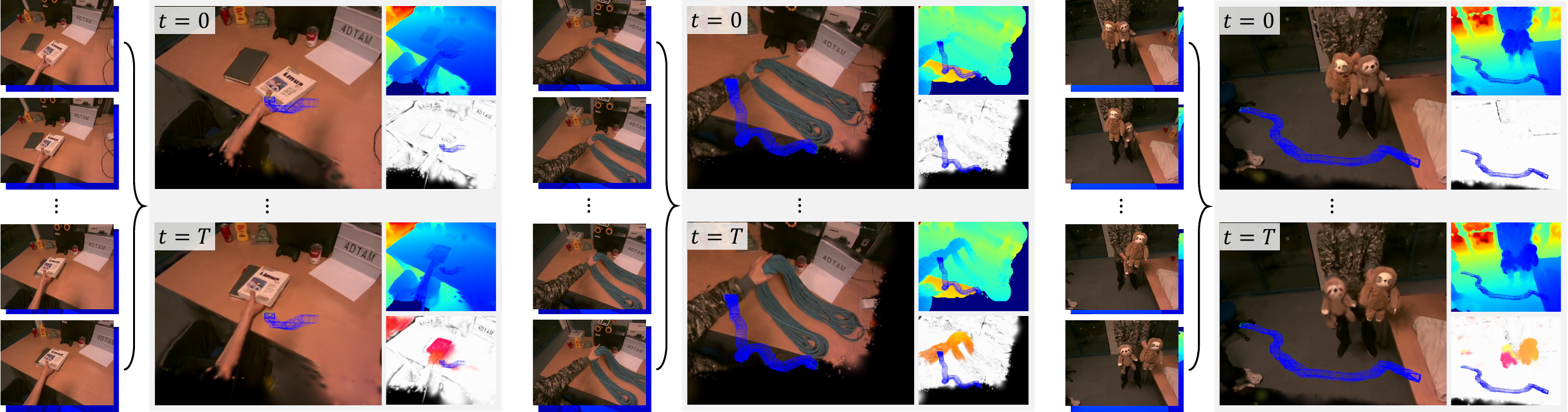}
\captionof{figure}{\textbf{4DTAM} jointly estimates camera-egomotion, appearance, geometry and scene dynamics without any template. \vspace{1em}}
\label{fig:teaser}
}]

\begin{abstract}
We propose the first 4D tracking and mapping method that jointly performs camera localization and non-rigid surface reconstruction via differentiable rendering. Our approach captures 4D scenes from an online stream of color images with depth measurements or predictions by jointly optimizing scene geometry, appearance, dynamics, and camera ego-motion. Although natural environments exhibit complex non-rigid motions, 4D-SLAM remains relatively underexplored due to its inherent challenges; even with 2.5D signals, the problem is ill-posed because of the high dimensionality of the optimization space. To overcome these challenges, we first introduce a SLAM method based on Gaussian surface primitives that leverages depth signals more effectively than 3D Gaussians, thereby achieving accurate surface reconstruction. To further model non-rigid deformations, we employ a warp-field represented by a multi-layer perceptron (MLP) and introduce a novel camera pose estimation technique along with surface regularization terms that facilitate spatio-temporal reconstruction. In addition to these algorithmic challenges, a significant hurdle in 4D SLAM research is the lack of reliable ground truth and evaluation protocols, primarily due to the difficulty of 4D capture using commodity sensors. To address this, we present a novel open synthetic dataset of everyday objects with diverse motions, leveraging large-scale object models and animation modeling. In summary, we open up the modern 4D-SLAM research by introducing a novel method and evaluation protocols grounded in modern vision and rendering techniques.
\end{abstract}    
\section{Introduction}
\label{sec:intro}

The world we live in has many moving elements. Rivers flow, trees sway, cookies crumble, and humans walk. 
Although Simultaneous Localization and Mapping (SLAM) methods which assume that most of the world is static are highly useful,
embodied agents which aim to navigate and interact with their environments in the most general way should be able to operate in dynamic scenes. There are several ways to segment and ignore moving scene elements, and a SLAM system can be assembled by integrating these individual modules so that it can reconstruct the static parts of a scene and estimate camera ego-motion. However, in this work, we aim for a more comprehensive spatio-temporal (4D) reconstruction of scenes exhibiting significant dynamic motion. Our primary focus is on a unified framework that leverages intrinsic capabilities of the underlying scene representation without heavily relying on prior assumptions about moving elements. 
\textbf{4D-SLAM} with general scene motion is difficult primarily because of the complex and high-dimensional nature of modeling non-rigid motions (and potential topological changes) while simultaneously optimizing the pose of a moving camera. 
There is much more redundancy than in rigid SLAM, and some prior assumptions are needed to combat this.
Another challenge lies in the lack of datasets to train and/or evaluate techniques.
Recent advances in computer vision and graphics make it a good time to revisit this problem. New 3D representations (e.g. neural fields and Gaussian splats) allow differentiable rendering of complex 3D scenes and optimization via 2D observations, and to model deformation fields smoothly without more specific assumptions. Also, the availability of high-quality 3D meshes on Internet and rendering software (e.g. Blender) gives the ability to render non-rigidly moving objects with ground truth.

We present \textbf{4DTAM}, a novel approach for \textbf{4D} \textbf{T}racking \textbf{A}nd \textbf{M}apping in dynamic scenes. We use Gaussian surface primitives to represent the scene and introduce a neural warp-field represented by a multi-layer perception (MLP) to model continuous temporal changes. We then utilize differentiable rendering to jointly optimize the scene geometry, appearance, dynamics, and camera ego-motion from an online stream of a single RGB-D camera. This enables accurate 3D reconstruction and real-time rendering, even in the presence of complex non-rigid deformations. To facilitate future research, we also introduce a new synthetic dataset of dynamic objects.  Our focus in this dataset is realistic, complex motion of scenes that are not well represented by existing deformable object models. Animated 3D meshes are rendered and the ground truth depth, surface normals, and foreground masks are extracted together with the camera poses/intrinsics. This dataset provides challenging scenarios for 4D reconstruction methods. We also release the full rendering script to allow the generation of custom 4D datasets. Our experimental results demonstrate that \textbf{4DTAM} achieves good performance in both camera tracking and scene reconstruction in the presence of dynamic objects. It can handle the complex motion of articulated objects (e.g., drawers) and non-rigid objects (e.g., curtains, flags, and animals), showcasing its potential for applications in robotics, augmented reality, and other fields requiring real-time dynamic scene understanding. We primarily use RGB-D sensor input, but also demonstrate an extension to monocular RGB streams by incorporating a monocular depth prediction network in the supplementary material.

In summary, the contributions of this paper are:

\begin{itemize} \item \textbf{4DTAM}, the first 4D tracking and mapping method that uses differentiable rendering and Gaussian surface primitives for dynamic environments. \item The first 2DGS~\cite{Huang2DGS2024}-based SLAM method with analytic camera pose gradients, normal initialization, and regularization to fully exploit depth signals. \item An MLP-based warp-field for modeling non-rigid scene, complemented by a novel camera localization technique and rigidity regularization of surface Gaussians. \item A novel 4D-SLAM dataset with complex object motions, ground-truth camera trajectories, and dynamic object meshes, along with an evaluation protocol. \item Extensive evaluations demonstrating that the method achieves state-of-the-art performance. \end{itemize}

\section{Related Work}
\label{sec:relatedworks}
\subsection{Visual SLAM}

Visual SLAM has been an extensively researched field, with Dense SLAM specifically focusing on capturing detailed scene geometry~\cite{Newcombe:etal:ICCV2011} and semantics~\cite{McCormac:etal:ICRA2017}. A central aspect of these methods lies in the choice of scene representation and the corresponding optimization framework. Dense SLAM methods based on traditional scene representations, such as volumetric Truncated Signed Distance Functions (TSDF)~\cite{Newcombe:etal:ISMAR2011, Whelan:etal:IJRR2015, InfiniTAM_ECCV_2016} or Surfels~\cite{Whelan:etal:RSS2015, badslam}, project 2D observations into 3D space and employ specific data fusion algorithms. While effective, these methods often fail to keep consistency between the model and sensor observations across multiple viewpoints, posing challenges for long-term operation. 

However, recent advancements in graphics hardware have facilitated the adoption of differentiable rendering frameworks, which have revolutionized inverse rendering and scene reconstruction~\cite{Kato:etal:CVPR2018, DVR,Mildenhall:etal:ECCV2020, mueller2022instant}. Differentiable rendering ensures multi-view consistency through streamlined backpropagation, enhancing scene reconstruction accuracy.
Notably, 3D Gaussian Splatting (3DGS)~\cite{kerbl3Dgaussians} has gained attention due to its flexible resource allocation and rapid forward rendering capabilities. Initially developed for photorealistic view synthesis, recent research has extended its application to surface reconstruction~\cite{guedon2023sugar, Yu2024GOF}. Enhanced methods, such as 2D Gaussian Splatting (2DGS)~\cite{Huang2DGS2024}, achieve superior geometry reconstruction by reducing the Gaussian dimension and explicitly defining surface normals. 
These differentiable rendering representations have been applied to visual SLAM, from coordinate-based MLPs~\cite{Sucar:etal:ICCV2021} to explicit voxel grids~\cite{Zhu2022CVPR, yang2022vox, johari-et-al-2023, wang2023coslam}, points~\cite{Sandström2023ICCV}, and 3D Gaussians~\cite{Matsuki:Murai:etal:CVPR2024, keetha2024splatam, yan2023gs}.

\subsection{SLAM for 4D Scene Reconstruction}

3D reconstruction of dynamic scenes has been extensively studied, with notable achievements using optimization methods, even for unknown non-rigid objects observed by a single moving RGB camera \cite{Torresani:etal:PAMI2008,Garg:etal:CVPR2013}. However, these approaches typically require batch optimization and are limited to smaller scenes. In contrast, dynamic SLAM targets incremental, reconstruction and tracking of large, continuously moving scenes ideally in real-time. Most methods to date have relied on RGB-D data from moving depth cameras.

While many methods detect and exclude dynamic objects to focus on static scene reconstruction \cite{Scona:etal:ICRA2018}, full spatiotemporal reconstruction (which we refer to as \textit{4D-SLAM}) requires more advanced solutions. For instance, tracking and reconstructing rigid moving objects separately \cite{Runz::Agapito::ICRA2017} or employing parametric shape models for known semantic classes like humans or animals \cite{SMPL:2015} are effective strategies. Specialized domains, such as endoscopic imaging, have utilized scene-specific priors or deformation models to handle non-rigid dynamics \cite{ma2021rnnslam,rodriguez2023nrslamnonrigidmonocularslam}.

An incremental 4D-SLAM for general dynamic scenes has remained more challenging, but has been addressed based on various regularizing assumptions and representations. DynamicFusion~\cite{Newcombe:etal:CVPR2015} pioneered a line of work~\cite{Slavcheva_2017_CVPR, Slavcheva_2018_CVPR, Innmann:etal:ECCV2016, gao18surfelwarp} which captures temporal evolution in the scene geometry by jointly optimizing a canonical volumetric representation (e.g., TSDF volume~\cite{Newcombe:etal:CVPR2015}) and a deformation field. As the solution space is extremely high-dimensional, additional constraints are often introduced to regularize the motion field~\cite{Slavcheva_2017_CVPR,Slavcheva_2018_CVPR} or to align visual features~\cite{Innmann:etal:ECCV2016,Bovzivc:etal:ARXIV2020}. Recent advances in 3D representations, such as neural fields and Gaussian primitives, have opened new possibilities for dynamic scene reconstruction. Canonical radiance and motion fields can be jointly optimized via differentiable rendering, as demonstrated with NeRF~\cite{park2021nerfies,pumarola2020d,tretschk2021nonrigid} and SDF~\cite{Cai2022NDR,wang2024morpheus}. For 3D Gaussians, which can explicitly represent points, motion can be estimated either through per-primitive trajectories~\cite{luiten2023dynamic} or learnable motion bases~\cite{som2024}. However, warp-field-based motion representation offers inherent smoothness regularization, leveraging the properties of neural fields~\cite{yang2023gs4d,yang2023deformable3dgs,deformgs,huang2023sc}. Most existing methods, however, rely on known camera poses or multi-camera setups to capture dense spatiotemporal observations. While DyNoMo~\cite{seidenschwarz2024dynomoonlinepointtracking} supports camera pose optimization, its 3D Gaussian representation is not suited for geometrically accurate reconstruction. In contrast, our 4DTAM framework enables 4D reconstruction using a single RGB-D camera, jointly optimizing camera poses, appearance, geometry, and dynamics, making it practical for most embodied agents.

\subsection{Datasets for 4D Reconstruction}

4D reconstruction has been studied extensively for the case of the human body. Datasets like Human3.6M~\cite{ionescu2013human3}, DeepCap~\cite{habermann2020deepcap}, and ZJU-MoCap~\cite{peng2021neural} capture diverse human motions under a multi-camera setup. The cameras are fixed, synchronized, and calibrated to reduce the difficulty in establishing dense multi-view correspondences. Only a small number of datasets provide single-stream RGB-D sequences captured from a moving camera~\cite{Slavcheva_2017_CVPR,bozic2020deepdeform,gao2022dynamic}. Recovering the camera poses is not trivial for such real-world captures, and additional post-processing (e.g. robust depth map alignment~\cite{wang2024morpheus}) is required. Another challenge lies in ground truth acquisition. Besides the depth measurements, other ground truths (e.g., scene flow, object mask) often require manual labeling. On the contrary, synthetic datasets~\cite{wang2020tartanair,Butler:etal:ECCV2012} provide perfect ground truths. Recent advances in open-source datasets~\cite{deitke2023objaverse} and rendering software~\cite{Blender} also close the synthetic-to-real domain gap significantly. To this end, we introduce a new high-quality synthetic dataset tailored for 4D reconstruction and camera pose estimation.

\begin{figure*}[!tbp]
  \center
    \includegraphics[width=\linewidth]{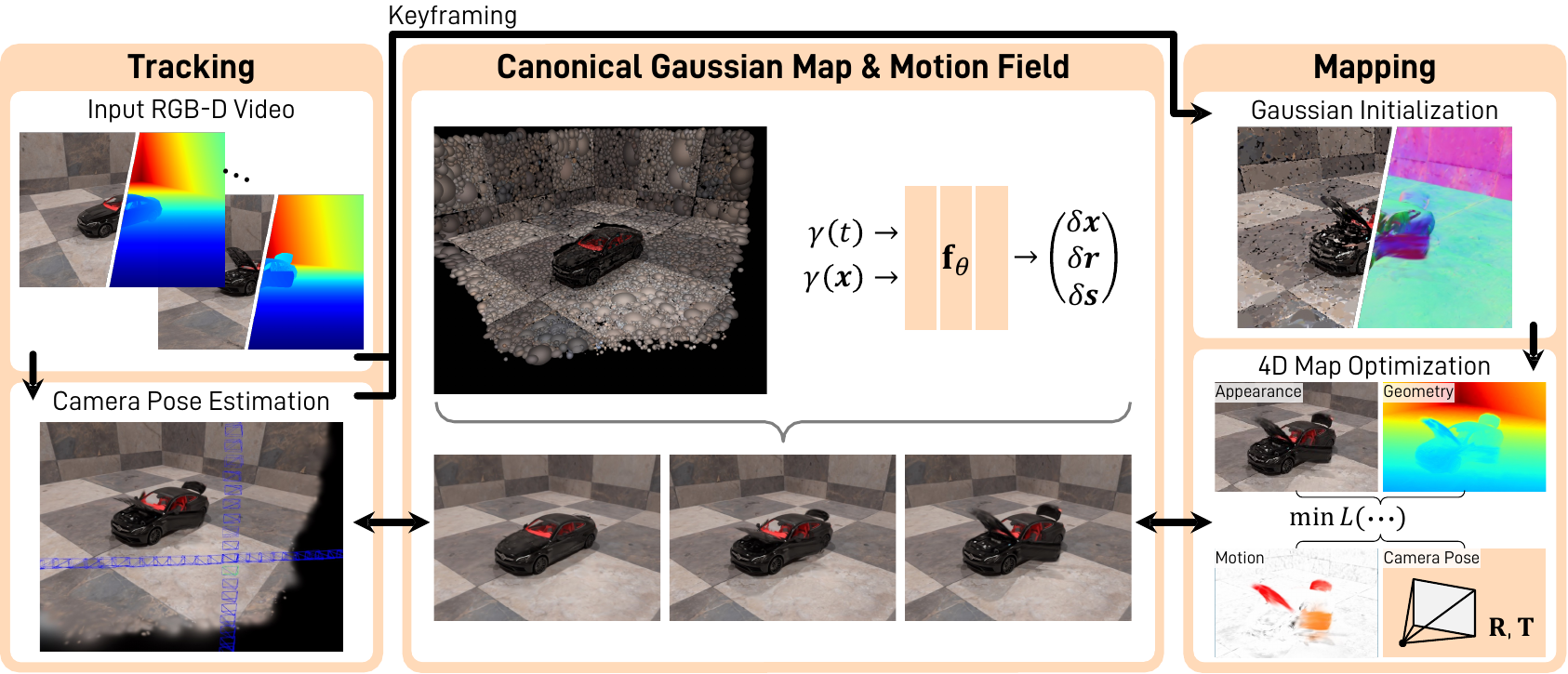}
  \caption{
   \textbf{Method overview of 4DTAM.}  
   }\label{fig:method}
\end{figure*}

\section{Method}
\subsection{2D Gaussian Splatting}
Our geometric scene representation is based on 2D Gaussian Splatting (2DGS)~\cite{Huang2DGS2024}. Unlike 3D Gaussian Splatting (3DGS), which uses blob-like splats, 2DGS functions as a stretchable surfel with explicitly defined surface normal directions. This property makes 2DGS particularly well-suited for non-rigid scene reconstruction with a single camera, where effectively handling 2.5D input signals is critical.

Each 2D Gaussian $\gaussians$ is represented by its 3D mean position $\mathbf{P}_{\mu}$, rotation $\mathbf{R}\in{\SO{3}}$, color $\mathbf{c}$, opacity $o$, and a scaling vector $\mathbf{S}\in \mathbb{R}^2$.  
The rotation matrix $\mathbf{R}$ is decomposed as $\mathbf{R} = [\mathbf{t}_u, \mathbf{t}_v, \mathbf{t}_w]$, where $\mathbf{t}_{u}$ and $\mathbf{t}_{v}$ represent two principal tangential vectors, and $\mathbf{t}_{w}$ is the normal vector, defined as $\mathbf{t}_{w}=\mathbf{t}_{u}\times \mathbf{t}_{v}$. For simplicity, spherical harmonics are omitted in this work.

The 2D Gaussian function is parameterized on the local tangent plane in world space as:
\begin{gather}
    P(u,v) = \bP_{\mu} + s_u \bt_u u + s_v \bt_v v = \bH(u,v,1,1)^{\mathrm{T}}\\
    \text{where} \, \bH = 
    \begin{bmatrix}
        s_u \bt_u & s_v \bt_v & \boldsymbol{0} & \bp_k \\
        0 & 0 & 0 & 1 \\
    \end{bmatrix} = \begin{bmatrix}
        \bR\bS & \bp_k \\ 
        \boldsymbol{0} & 1\\
    \end{bmatrix}
\end{gather}

For a point $\mathbf{u} = (u, v)$ in the tangential plane of 2D Gaussian ($uv$ space), its projection onto the image plane is given by 
\begin{equation}
    \bx = (xz, yz, z, 1)^\mathrm{T} = \bW \mathbf{P}(u,v) = \bW \bH (u,v,1,1)^\mathrm{T}
\label{eq:projection}
\end{equation}
where $\bW \in \mathbb{R}^{4\times 4}$ is the transformation matrix from world space to screen space.

To avoid numerically unstable matrix inversion of $\mathbf{M} = (\mathbf{W}\mathbf{H})^{-1}$, 2DGS applies ray-splat intersection by finding the intersection of non-parallel planes (x-plane and y-plane). The ray $\bx = (x, y)$ is determined by the intersection of the $x$-plane $\bh_x$ and the $y$-plane $\bh_y$, represented as {$\bh_x = (-1, 0, 0, x)^{\mathrm{T}}$} and {$\bh_y = (0, -1, 0, y)^{\mathrm{T}}$}, respectively. In the uv coordinates of the 2D Gaussian, this is expressed as:
\begin{equation}
\bh_u = (\bW \bH)^\mathrm{T} \bh_x \quad \text{and} \quad \bh_v = (\bW \bH)^\mathrm{T} \bh_y
\label{eq:plane_transformation}
\end{equation}
The intersection point meets the following condition,
\begin{equation}
    \bh_u \cdot (u,v,1,1)^\mathrm{T} = \bh_v \cdot (u,v,1,1)^\mathrm{T} = 0
\end{equation}
This leads to an solution for the intersection point $\bu(\bx)$:
\begin{equation}
u(\bx) = \frac{\bh_u^2 \bh_v^4 - \bh_u^4 \bh_v^2}{\bh_u^1 \bh_v^2-\bh_u^2 \bh_v^1} \qquad v(\bx) = \frac{\bh_u^4 \bh_v^1 - \bh_u^1 \bh_v^4}{\bh_u^1 \bh_v^2-\bh_u^2\bh_v^1}
\label{eq:uv_intersection}
\end{equation} where $\bh_u^i,\bh_v^i$ are the $i$-th parameter of the 4D homogeneous plane parameters. 

The 2D Gaussian at $(u, v)$ is evaluated as:
\begin{equation}
\mathcal{G}(\mathbf{u}) = \exp\left(-\frac{u^{2} + v^{2}}{2}\right)
\end{equation}

The 2D Gaussians are sorted along the camera ray by their center depth and organized into image tiles. Per-pixel color is rendered via volumetric alpha blending:

\begin{align}
c(\mathbf{x}) &= \sum_{i=1} \mathbf{c}_{i} \alpha_{i} \mathcal{G}_i(\mathbf{u}(x)) \prod_{j=1}^{i-1} (1 - \alpha_{j} \mathcal{G}_j(\mathbf{u}(x)))
\end{align}

where depth and normal can be rendered similarly.

\subsection{Analytic Camera Pose Jacobian}
\label{subsec:jacobian}
One major advantage of Gaussian Splatting is its analytical formulation of gradient flow for model parameters, enabling real-time full-resolution rendering. However, it assumes posed images as input and does not provide gradients for camera poses. To accelerate optimization, we derive the analytic Jacobian of the camera pose for 2D Gaussian Splatting and implement it using a CUDA kernel. This formulation has potential applications for a wide range of tasks involving pose estimation in surface-based Gaussian Splatting.

We use Lie algebra to derive the minimal Jacobians for the camera pose matrix from the world coordinate system to the camera's local coordinate system, defining $\camCW \in \SE{3}$ and $\tauC \in \se{3}$. Since 2DGS backpropagates gradients to $\bM^{T} = \bW\bH$ during the optimization of the 3D mean, we require the partial derivative $\pd{\bM^{T}}{\tauC}$. Let $\mathbf{K} \in \mathbb{R}^{4 \times 4}$ represent the camera projection matrix. Then, equation~\ref{eq:projection} is rewritten as:
\begin{equation}
    \bx = \bM^{T}(u,v,1,1)^\mathrm{T} = \bK \camCW \bH (u,v,1,1)^\mathrm{T}
\label{eq:projection_Tcw}
\end{equation}

Using the chain rule, the partial derivatives are computed as:
\begin{align}
    \pd{\bM^{T}}{\tauC} &= \pd{\bM^{T}}{\bW}\pd{\bW}{\camCW}\pd{\camCW}{\tauC}, \\
    \pd{\camCW}{\tauC} &= \begin{bmatrix}
    \mathbf{0} & -{\rotCW}_{:, 1}^\times \\
    \mathbf{0} & -{\rotCW}_{:, 2}^\times \\
    \mathbf{0} & -{\rotCW}_{:, 3}^\times \\
    \identity & -{\bt_{CW}}^\times
    \end{bmatrix}
\label{eqn:grad_tcw}
\end{align}

where ${\rotCW} \in \SO{3}$ and ${\bt}_{CW} \in \mathbb{R}^{3}$ denote the rotation and translation parts of $\camCW$, respectively. The notation ${}^{\times}$ represents the skew-symmetric matrix of a 3D vector, and ${\rotCW}_{:, i}$ denotes the $i$th column of ${\rotCW}$.

2DGS also renders a normal map, which can be supervised using the loss computed from the rendered normals. Let $\bn_c$ denote the camera-space normal. The normal of a 2D Gaussian in the camera's local coordinate system is defined as:
\begin{align}
    \bn_c &= \camCW \bt_w
\label{eqn:world_normal}
\end{align}
where $\bt_w$ is the surface normal in the world coordinate system.

Borrowing the notation of the left Jacobian for Lie groups from~\cite{Sola:etal:ARXIV2018}, the partial derivative is given by:
\begin{align}
    \pd{\bn_c}{\tauC} = \mpd{\bn_c}{\camCW} =  \begin{bmatrix}\identity & -\bn_c^\times\end{bmatrix}
\label{eqn:grad_normal}
\end{align}

Further details of the derivation are provided in the supplementary material.

\subsection{Warp Field}
To model time-varying deformations, we use a warp-field represented by a coordinate-based network~\cite{yang2023deformable3dgs, yang2023gs4d}. In our hand-held single-camera setup, the limited view coverage of dynamic objects necessitates structural priors in the motion representation. For this, we employ a compact MLP as the warp-field to estimate transitions from the canonical Gaussians following~\cite{yang2023deformable3dgs}.

\par Given time $t$ and center position $\boldsymbol{x}$ of 2D Gaussians in canonical space as inputs, the deformation MLP $\mathbf{f_\theta}$ produces offsets, which subsequently transform the canonical 2D Gaussians to the deformed space:
\begin{equation}
    (\delta \boldsymbol{x}, \delta \boldsymbol{r}, \delta \boldsymbol{s}) = \mathbf{f_\theta}(\gamma_{1}(\boldsymbol{x}), \gamma_{2}(t))
\end{equation} 
where $\delta \boldsymbol{x}\in\mathbb{R}^{3}$, $\delta \boldsymbol{r}\in\SO{3}$, $\delta \boldsymbol{s} \in\mathbb{R}^{2}$ denotes the offsets of 2D Gaussian's mean position, rotation and scale respectively, $\gamma$ denotes the frequency-based positional encoding ~\cite{Mildenhall:etal:ECCV2020}. For deformable SLAM applications, we leverage a CUDA-optimized MLP implementation~\cite{mueller2022instant} to enable fast, interactive reconstruction.

\subsection{Tracking and Mapping Framework}
Our SLAM method follows the standard tracking and mapping architecture, where the tracking module is in charge of fast online camera pose estimation while the mapping performs a relatively more involved  joint opimtization of the camera poses, geometry and motion of selected keyframes. Further details of the hyperparameters are available in the supplementary material. 

\subsubsection{Tracking}
The tracking module estimates the coarse camera pose for the latest incoming frame. This is achieved by minimizing the photometric and depth rendering errors between the sensor observation and the rendering from the deformable Gaussian model. Unlike static 3DGS SLAM methods, we estimate the camera pose relative to the warped Gaussians at the latest keyframes timestamp $t_{kf}$, assuming the deformed scene structure at $t_{kf}$ is closest to the current state.
We define photometric rendering loss as:
\begin{equation}
    L_{p} = \left\| I(\gaussians_{cano}, \camCW, t_{kf}) - \gtimage \right\|_1~
    \label{eqn:photometric}
\end{equation}
Here $I(\gaussians, \camCW)$ denotes a rendered color image from the cannonical Gaussians $\gaussians_{cano}$, timestamp of the latest keyframe $t_{kf}$ and camera pose $\camCW$ , and $\gtimage$ is an observed image. Similarly, we also minimize geometric depth error:
\begin{equation}
    L_{g} = \left\| D(\gaussians_{cano}, \camCW, t_{kf}) - \gtdepth \right\|_1
    \label{eqn:geometric_residual}
\end{equation}

Following MonoGS~\cite{Matsuki:Murai:etal:CVPR2024}, we further optimize affine brightness parameters. Keyframes are selected every N-th frame and sent to the mapping process for further refinement.

\subsubsection{Mapping}

The mapping module performs joint optimization of the camera pose, canonical Gaussians, and the warp field within a sliding window.

\paragraph{Gaussian Management}
\begin{figure}[!tbp]
  \center
  \includegraphics[width=\linewidth]{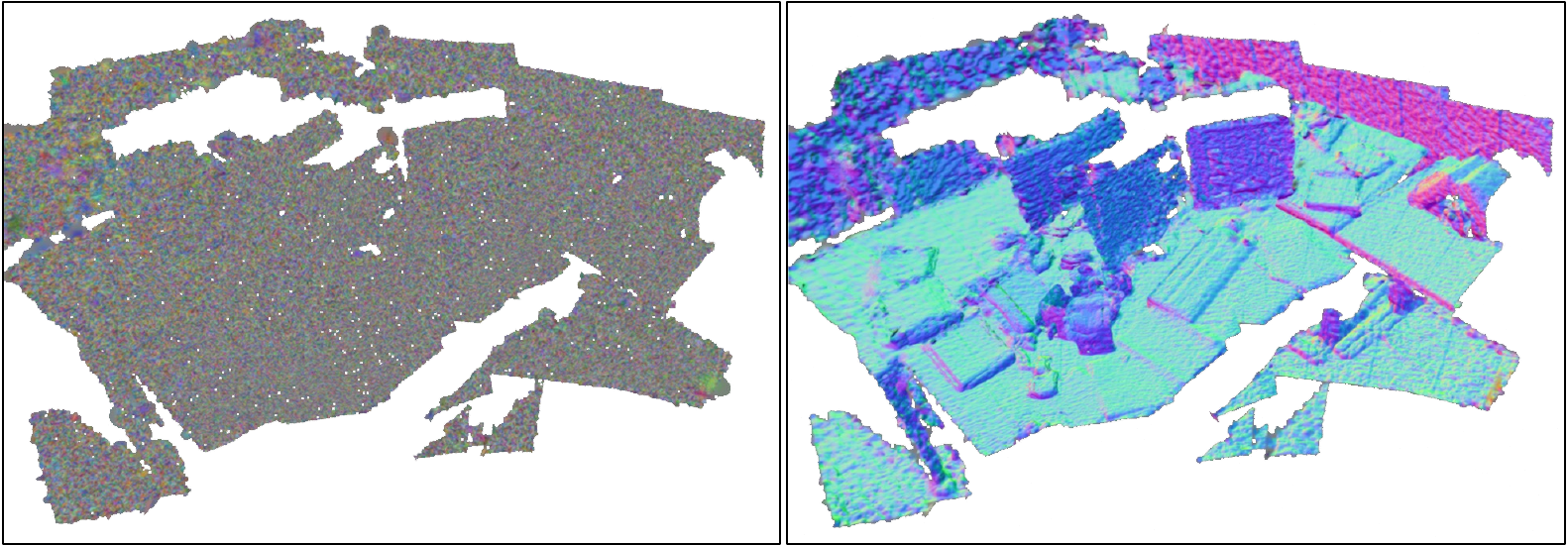}
  \caption{   \textbf{2D Gaussian's Surface Normal Rendering based on Different Initialization.} Left: Random initialization. Right: Our initialization aligned with sensor measurement.}\label{fig:2dgs_init}
\end{figure}

When a new keyframe is registered, we add new Gaussians to the canonical Gaussians 
$\gaussians_{cano}$, based on the back-projected point cloud from the RGB-D observations. Unlike 3DGS, 2DGS explicitly encodes surface normal information in its rotation vector, making it beneficial to initialize using surface normals estimated from sensor depth measurements. To achieve this, we compute the surface normals of the current depth observation by taking the finite difference of neighboring back-projected depth points and assign them as the normal vectors of the 2D Gaussianss $\bt_{w}$. This is formulated as:
\begin{equation}
\bt_{w} = \frac{\nabla_x \bp_d \times \nabla_y \bp_d}{|\nabla_x \bp_d \times \nabla_y \bp_d|}
\label{eq:normal_depth}
\end{equation}
where $\bp_d$ denotes points back-projected by the current sensor depth observation. We store the computed normal information as a 2D image $\bN_{sensor}$ for normal supervision. Pruning and densification parameters follow MonoGS, which effectively prunes the wrongly inserted Gaussians in the canonical space due to the object movement.

\paragraph{4D Map optimization} 
We perform joint optimization of the camera ego-motion, appearance, geometry and scene dynamics. In a single-camera setup, the lack of spatiotemporally dense observations makes fully capturing dynamic scenes challenging, as complete spatial (xyz) coverage over time (t) is only feasible with multi-camera systems. To address this, we introduce regularization terms for both shape and motion.

In addition to photometric and depth losses, we apply a normal regularization based on sensor measurements to better align 2D Gaussians. Unlike the original 2DGS methods, which compute normals by finite differences of rendered depth during every optimization step—leading to high computational costs—we instead propose to use normals precomputed from depth input as supervision. This reduces computational overhead, as normals are calculated only when a new keyframe is inserted:
\begin{equation}
L_{n} = \sum_{i\in{h{\times}{w}}} (1-\bn_i^\mathrm{T}\bN_{sensor,i})
\end{equation}

To constrain motion in unobserved regions, we apply an as-rigid-as possible regularization loss $L_{ARAP}$ from ~\cite{luiten2023dynamic} to the Gaussian means. Additionally, we introduce a novel surface normal rigidity loss, constraining the 2D Gaussians' surface normals to stay similar between timesteps $t_{1}$ and $t_{2}$, preserving local surface rigidity:
\begin{equation}
L_{ARAP\_n} = w_{i,j}\left\| ({\bt_w})_{i, t_1}^{T}({\bt_w})_{j, t_1} -  ({\bt_w})_{i, t_2}^{T}({\bt_w})_{j, t_2}  \right\|_1
\label{eq:normal_rigidity}
\end{equation}
where $w_{i,j}$ is a distance-based weighting factor like $L_{ARAP}$.
We apply ARAP regularizers between the oldest and latest keyframe in the current window.

Together with the isotropic loss $L_{iso}$ proposed in~\cite{Matsuki:Murai:etal:CVPR2024}, we minimize the following total cost function:
\begin{align}
L_{total} &= \lambda_{p}L_{p} + \lambda_{g}L_{g} + \lambda_{n}L_{n} \notag \\
&\quad + \lambda_{iso}L_{iso} + L_{ARAP} + L_{{ARAP}\_n}
\end{align}

The optimization is based on the sliding window heuristics in ~\cite{Engel:etal:PAMI2017}, with two additional keyframes randomly selected from the history.

\paragraph{\textbf{Global Optimization}}
Sliding window-based optimization prioritizes the latest frame, causing past keyframe information to degrade over time. After tracking, if required we can perform global optimization to finalize the map, which takes less than 1 minute on an RTX 4090. During this step, the poses and number of Gaussians are fixed, and one keyframe is randomly selected per iteration. The process uses the normal consistency loss of 2DGS, ensuring global consistency despite being relatively slow.

\subsection{Dataset Generation}

\begin{figure*}[!tbp]
  \center
  \includegraphics[width=\linewidth]{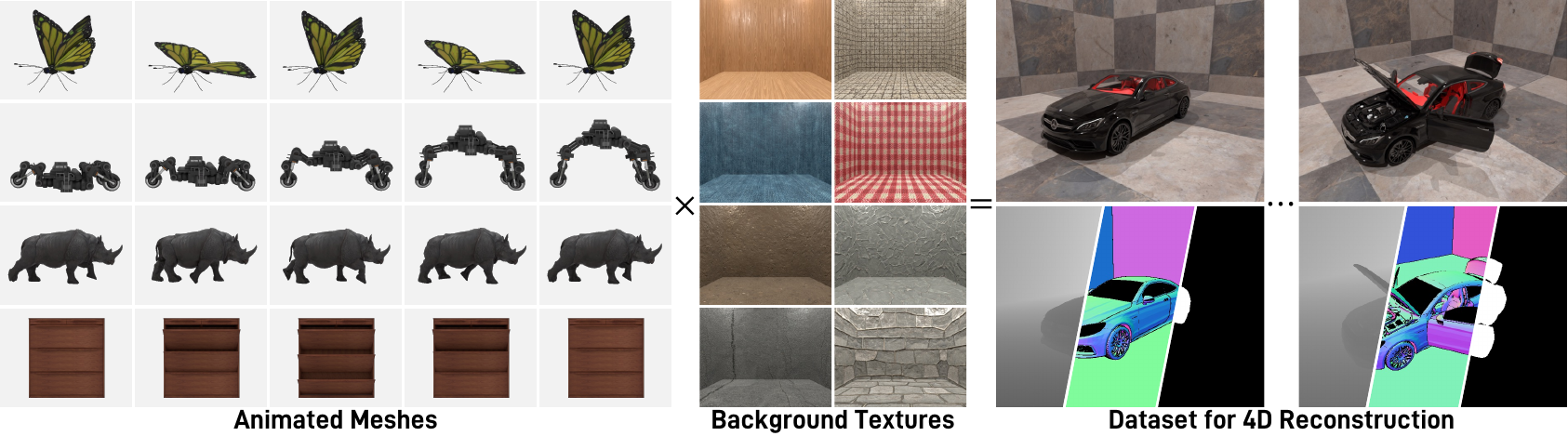}
  \caption{
   \textbf{Sim4D dataset.} We create a new dataset for 4D reconstruction by rendering animated 3D meshes. 
   }\label{fig:sim4d}
\end{figure*}


We introduce \textbf{Sim4D}, a new synthetic dataset for 4D reconstruction. Recently, a large number of photo-realistic, animated 3D meshes have become available~\cite{deitke2023objaverse,Sketchfab}. Combined with open-source graphics software~\cite{Blender,blendersynth}, such meshes provide a scalable way of generating datasets for non-rigid 4D reconstruction. The data generation pipeline is illustrated in Fig.~\ref{fig:sim4d}. 

\noindent
\textbf{Meshes and background.} We collected 50 high-quality, animated 3D meshes from Objaverse~\cite{deitke2023objaverse} and Sketchfab~\cite{Sketchfab}, all of which are under CC-BY license. The collected meshes exhibit a wide variety of motions, including non-rigid deformation and topological changes. We then place the object inside a cube and randomize the background texture. Texture maps are collected from Poly Haven~\cite{PolyHaven} and are all under CC0 license.

\noindent
\textbf{Rendering.} We render 240 to 540 frames for each object. The camera trajectories are defined along arcs of 20 degrees, and test viewpoints are defined outside of these arcs to evaluate the performance of novel-view synthesis and to quantify the accuracy of the reconstructed geometry. At each timestamp, the RGB image, ground truth depth, surface normals, and foreground mask are rendered and the camera intrinsics/extrinsics saved. Please refer to the supplementary material for additional details.
\section{Evaluation}
\subsection{Experimental Setup}
We extensively evaluate our non-rigid SLAM method on both synthetic and real-world datasets. Previous non-rigid RGB-D SLAM work has primarily focused on qualitative demonstrations using limited datasets, showcasing the early-stage potential of the field. To advance research, we introduce a quantitative evaluation protocol with the new Sim4D dataset. Our evaluation covers camera pose accuracy, as well as the appearance and geometric quality of the reconstructed models. Additionally, we demonstrate real-world performance using a self-captured dataset.

While designed primarily for dynamic scenes, our method is the first to leverage surface Gaussian splatting for both static SLAM and non-rigid RGB-D reconstruction. To further validate our approach, we perform a detailed quantitative component-wise ablation analysis.

\paragraph{Metrics and Datasets}  
For our main {\bf Non-Rigid SLAM evaluation}, we evaluate our method on 8 sequences from the Sim4D dataset. We first report ATE RMSE for trajectory evaluation. To assess SLAM map quality, we report depth rendering error (L1 error) for geometry and PSNR, SSIM, and LPIPS for appearance evaluation. For Sim4D, metrics are calculated from test views (extrapolated positions across different timestamps). The estimated and ground truth trajectories are aligned on the first frame, and test view positions are queried in the ground truth trajectory's coordinate system. Details about the test viewpoints are in  the supplementary material. Since SurfelWarp~\cite{gao18surfelwarp} requires explicit foreground segmentation, we collect its results only on pixels with valid reconstruction. For {\bf Static SLAM ablation}, we report ATE RMSE, rendering performance, and TSDF-fused mesh metrics, following the protocol in ~\cite{Sandström2023ICCV}. We evaluate our method on the Replica~\cite{replica19arxiv} dataset and the TUM RGB-D dataset~\cite{Sturm:etal:IROS2012}. To isolate the impact of scene representation from system differences, we replaced MonoGS's representation with 2DGS while keeping all other system configurations identical. For {\bf Offline Non-Rigid Reconstruction ablation}, we report the average geometry and appearance rendering metrics on subsets of the DeepDeform~\cite{bozic2020deepdeform}, KillingFusion~\cite{Slavcheva_2017_CVPR}, and iPhone datasets~\cite{gao2022dynamic}, which are used in~\cite{wang2024morpheus}. Numerical quantities for each sequence is available in supplementary material. Since ~\cite{wang2024morpheus} primarily focuses on object shape completion, metrics are calculated only within the given segmentation mask. The camera pose is provided by the dataset, and pose optimization is disabled to focus solely on reconstruction performance. We perform 30000 iteration for training, which takes approximately 30 mins.

\paragraph{Baseline Methods}  
For quantitative non-rigid SLAM evaluation, we compare our method with SurfelWarp~\cite{gao18surfelwarp}, the only non-rigid RGB-D SLAM method with publicly available code. For component-wise ablation analysis, we compare against MonoGS~\cite{Matsuki:Murai:etal:CVPR2024} for static SLAM evaluation and Morpheus~\cite{wang2024morpheus} for offline reconstruction.

\paragraph{Implementation Details}  
Our SLAM system runs on a desktop equipped with an Intel Core i9-12900K (3.50GHz) processor and a single NVIDIA GeForce RTX 4090 GPU. The camera pose jacobian for 2DGS, described in Section~\ref{subsec:jacobian}, is implemented using a CUDA rasterizer, similar to other gradients in Gaussian Splatting. For real-world data capture, we used the Realsense D455.

\begin{table*}[h!]
\centering
\resizebox{\textwidth}{!}{%
\begin{tabular}{c|>{\centering\arraybackslash}m{2.0cm}|l|cccccccc}
\hline
Method      & Category        & Metric            & curtain & flag & mercedes & modular\_vehicle & rhino & shoe\_rack & water\_effect & wave\_toy \\ \hline
\multirow{5}{*}{SurfelWarp~\cite{gao18surfelwarp}}
& \multirow{1}{*}{\centering Trajectory} & ATE RMSE[cm]↓& 6.10 & 31.9 & 5.21 & 4.21 & 2.81 & 2.16 & 2.60 & 1.45 \\ \cline{2-11}
& \multirow{1}{*}{\centering Geometry} & L1 Depth[cm]↓& 49.1 & 50.8 & 5.2 & 10.3 & 44.9 & 4.25 & 46.7 & 47.5 \\
\cline{2-11}
& \multirow{3}{*}{\centering Appearance} & PSNR [dB] ↑ & 15.78 & 11.04 & 25.7 & 19.45 & 16.76 & 26.3 & 17.3  & 16.4 \\
&                            & SSIM ↑            & 0.468 & 0.343 & 0.779 & 0.362 & 0.188 & 0.795 & 0.325 & 0.364 \\
&                            & LPIPS ↓           & 0.56 & 0.659 & 0.483 & 0.638 & 0.665 & 0.397 & 0.587 & 0.555 \\ \hline
\multirow{5}{*}{Ours}
& \multirow{1}{*}{\centering Trajectory} & ATE RMSE[cm]↓ & \bf{0.25} & \bf{1.00} & \bf{0.18} & \bf{0.31} & \bf{0.25} & \bf{0.18} & \bf{0.29} & \bf{0.32} \\ \cline{2-11}
& \multirow{1}{*}{\centering Geometry} & L1 Depth[cm]↓& \bf0.96 & \bf3.58 & \bf0.62 & \bf1.44 & \bf1.85 & \bf0.99 & \bf3.20 & \bf3.43 \\
\cline{2-11}
& \multirow{3}{*}{\centering Appearance} & PSNR [dB] ↑
& \bf28.01 & \bf21.01 & \bf32.13 & \bf30.59 & \bf24.13 & \bf31.7 & \bf27.12 & \bf27.10 \\
&                            & SSIM ↑            & \bf0.787 & \bf0.601 & \bf0.894 & \bf0.801 & \bf0.742 & \bf0.901 & \bf0.795 & \bf0.794 \\
&                            & LPIPS ↓           & \bf0.096 & \bf0.150 & \bf0.138 & \bf0.210 & \bf0.260 &\bf 0.12 & \bf0.908 & \bf0.097 \\ \hline
\end{tabular}
}
\caption{\textbf{Non-rigid SLAM Evaluation on Sim4D Dataset.}}
\label{table:slam_sim4d}
\end{table*}

\subsection{Quantitative Evaluation}  
\begin{figure}[!tbp]
  \center
  \includegraphics[width=\linewidth]{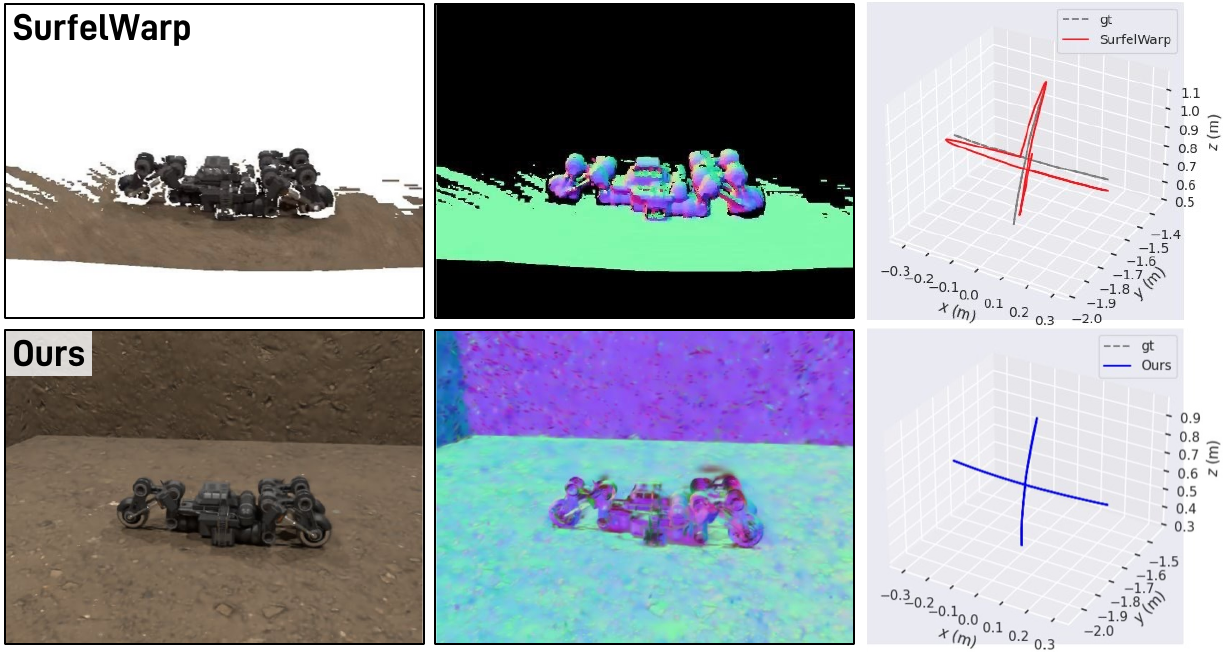}
  \caption{
   \textbf{Qualitative comparison to SurfelWarp.} Left: Rendered image, Middle: Rendered normal map, Right: Estimated camera trajectory}\label{fig:surfelwarp}
\end{figure}

Table~\ref{table:slam_sim4d} compares our method with SurfelWarp~\cite{gao18surfelwarp}. Our method outperforms SurfelWarp across all metrics. To analyze this further, Fig.~\ref{fig:surfelwarp} provides qualitative visualizations and trajectory plots for the modular\_vehicle sequence. Since SurfelWarp relies on a foreground mask, its reconstruction lacks scene completeness. In contrast, our method reconstructs the entire scene within a joint optimization framework, providing more comprehensive coverage. Additionally, compared to SurfelWarp's back-projection and Surfel fusion scheme, our differentiable rendering-based optimization enforces multi-view consistency over time, resulting in superior camera tracking and consistent 3D reconstruction. Our method achieves camera pose estimation at approximately 1.5 fps and completes the final global optimization in 1 minute. 

\subsection{Qualitative Evaluation}
\begin{figure}[!tbp]
  \center
  \includegraphics[width=\linewidth]{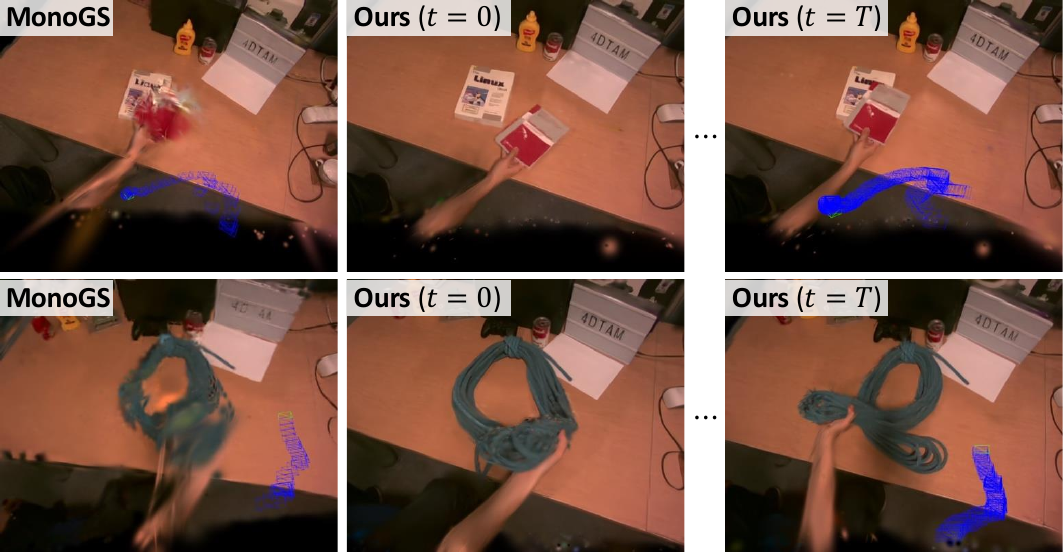}
  \caption{
   \textbf{Qualitative Results on Real-World Datset.} Our method effectively handles dynamic objects compared to MonoGS.
   }\label{fig:qualitative}
\end{figure}

Fig.~\ref{fig:qualitative} presents qualitative reconstruction results on real-world dynamic scenes. Our method successfully reconstructs dynamic scenes with non-rigid deformations, whereas MonoGS fails to handle such complexities.

\subsection{Ablation Study}
\paragraph{Static SLAM}

\begin{figure}[!tbp]
  \center
  \includegraphics[width=\linewidth]{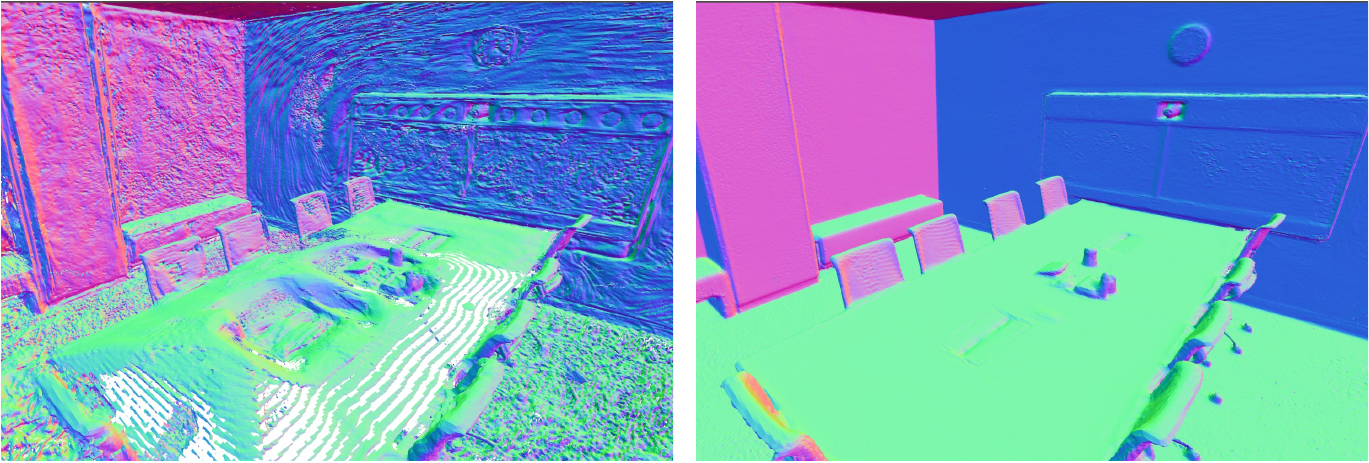}
  \caption{
   \textbf{3D Reconstruction Result on Replica Office4.} Left: MonoGS. Right: Ours (MonoGS-2D). Our surface Gaussian-based approach yields more accurate geometric reconstructions.
   }\label{fig:replica_mesh}.
\end{figure}

\begin{table}[h!]
\centering
\resizebox{\columnwidth}{!}{%
\begin{tabular}{l|c|ccccccccc}
\hline
 & Metric        & r0    & r1    & r2    & o0    & o1    & o2    & o3    & o4& avg      \\ \hline
\multirow{5}{*}{MonoGS} 
                        & ATE RMSE[cm]↓ & 0.44  & \bf0.32  & \bf0.31  & 0.44  & 0.52  & 0.23  & \bf0.17  & 2.25  &0.59   \\ \cline{2-11}
                        & Depth L1[cm]↓   & 3.00  & 3.47  & 4.66  & 3.10  & 6.08  & 6.15  & 4.77  & 4.94 &4.52   \\
                        & Precision[\%]↑  & 39.0  & 28.8  & 28.9  & 39.7  & 15.8  & 28.0  & 32.5  & 25.5 &29.7   \\
                        & Recall[\%]↑     & 44.2  & 34.5  & 32.8  & 47.6  & 24.3  & 30.0  & 35.4  & 28.5 &34.6   \\
                        & F1[\%] ↑         & 41.5  & 31.4  & 30.7  & 43.3  & 19.1  & 29.0  & 33.9  & 26.9 &31.9   \\ \hline
\multirow{5}{*}{\bf{MonoGS-2D}}   
                        & ATE RMSE[cm]↓ & \bf0.42  & 0.43  & 0.35  & \bf0.19  & \bf0.19  & \bf0.22  & 0.27  & \bf0.80 & \bf0.36   \\ \cline{2-11}
                        & Depth L1[cm]↓   & \bf0.45  & \bf0.28  & \bf0.57  & \bf0.37  & \bf0.59  & \bf0.85  & \bf0.62  &  \bf0.63 &\bf0.54   \\
                        & Precision[\%]↑  & \bf97.0  & \bf97.0  & \bf97.0  & \bf97.1  & \bf97.9  & \bf95.8  & \bf94.8  & \bf83.9 &\bf95.0   \\
                        & Recall[\%]↑     & \bf85.5  & \bf86.0  & \bf84.8  & \bf89.4  & \bf85.1  & \bf81.8  & \bf81.5  & \bf72.4 &\bf83.3   \\
                        & F1[\%]↑         & \bf90.9  & \bf91.3  & \bf90.5  & \bf93.1  & \bf91.1  & \bf88.2  & \bf87.6  & \bf77.7 &\bf88.8   \\ \hline
\end{tabular}
}
\caption{\textbf{Static SLAM Ablation on Replica.}}
\label{table:static_replica}

\end{table}

\begin{table}[h!]
\centering
\resizebox{\columnwidth}{!}{%
\begin{tabular}{c|l|cccc}
\hline
Method      & Metric        & fr1/desk & fr2/xyz & fr3/office & avg. \\ \hline
\multirow{2}{*}{MonoGS} 
            & ATE RMSE[cm]↓ & \bf{1.50} & 1.44 & \bf1.49 & \bf{1.47}  \\ 
            & Depth L1[cm]↓   & 6.2     &  13.0 & 13.0 & 10.7    \\ \hline
\multirow{2}{*}{MonoGS-2D}     
            & ATE RMSE[cm]↓ & 1.58     & \bf{1.20} & 1.83 & 1.57   \\ 
            & Depth L1[cm]↓   & \bf{3.00}  & \bf{2.30}   & \bf{4.30}  &  \bf{3.2}   \\ \hline
\end{tabular}%
}
\caption{\textbf{Static SLAM Ablation on TUM}}
\label{table:static_tum}
\end{table}

\begin{table}[h!]
\centering
\resizebox{\columnwidth}{!}{%
\begin{tabular}{c|l|ccc}
\hline
Method      & Metric        & KillingFusion & DeepDeform & iPhone \\ \hline
\multirow{4}{*}{Morpheus~\cite{wang2024morpheus}} 
            & Depth L1[cm]↓ & \bf{3.2}           & 1.9        & 2.4    \\ \cline{2-5}
            & PSNR [dB] ↑   & 27.02         & \bf{26.81}         & \bf{25.28}     \\
            & SSIM ↑        & 0.77          & 0.81       & 0.46   \\
            & LPIPS ↓       & 0.40          & 0.38       & 0.63   \\ \hline
\multirow{4}{*}{Ours}     
            & Depth L1[cm]↓ & 4.9           & \bf{1.1}        & \bf{0.57}   \\ \cline{2-5}
            & PSNR [dB] ↑   & \bf{31.13}            & 24.15         & 27.54     \\
            & SSIM ↑        & \bf{0.93}          & \bf{0.90}       & \bf{0.79}   \\
            & LPIPS ↓       & \bf{0.13}          & \bf{0.27}       & \bf{0.26}   \\ \hline
\end{tabular}%
}
\caption{\textbf{Offline Non-Rigid Reconstruction Ablation:} Rendering Error Metrics on Real-world Dataset.}
\label{table:offline_nonrigid}
\end{table}

Table~\ref{table:static_replica} provides the camera ATE and 3D reconstruction evaluation results.
Our 2DGS-based implementation shows competitive performance and achieves the best result in 6 out of 8 sequences for camera ATE, and consistently better result on rendering and 3D reconstruction metrics. The reconstruction is visualized in Fig~\ref{fig:replica_mesh} which shows the comparison of the mesh generated by TSDF Fusion between MonoGS and MonoGS-2D. Table~\ref{table:static_tum} provides the camera ATE and rendering metrics evaluation on TUM dataset. Our method shows on par camera ATE but shows the increased geometric reconstruction quality.

\begin{figure}[!tbp]
  \center
  \includegraphics[width=\linewidth]{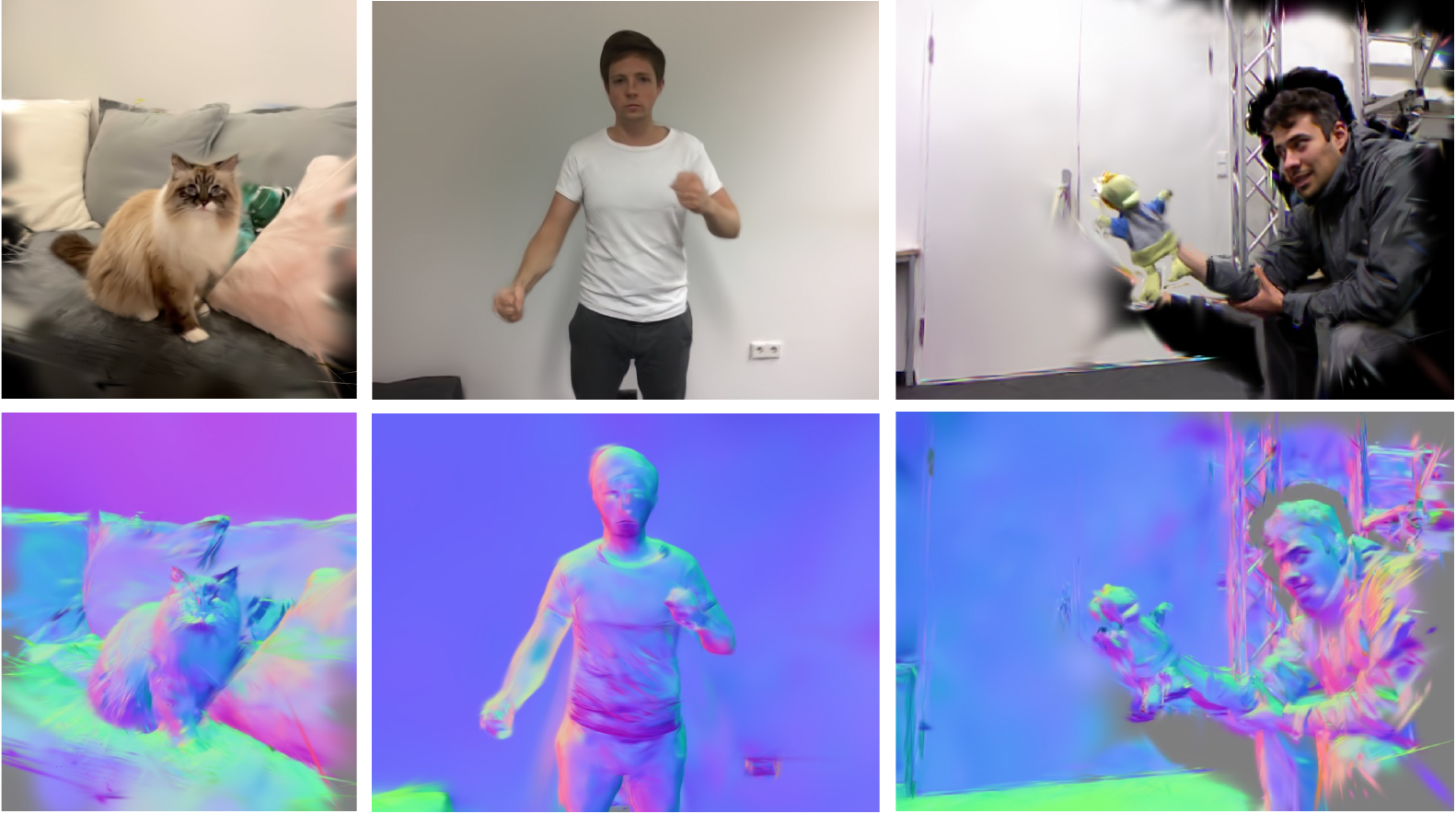}
  \caption{
   \textbf{Non-rigid Reconstruction Results.} Our method flexibly models non-rigid deformations without requiring any shape templates or foreground/background separation.
   }\label{fig:offline_nonrigid}
\end{figure}

\paragraph{Offline Non-rigid RGB-D Surface Reconstruction}
Table~\ref{table:offline_nonrigid} reports offline reconstruction results, where camera poses are given.  Our 2DGS+MLP deformation model shows competitive rendering performance compared to NeRF based methods. Note that Gaussian Splatting has the additional advantage of its rendering speed. We further provide qualitative visualizations in Fig.~\ref{fig:offline_nonrigid}.

\section{Conclusion}
We presented the first tracking and mapping method for non-rigid surface reconstruction using Surface Gaussian Splatting. Our approach integrates a 2DGS + MLP warp-field SLAM framework with camera pose estimation and regularization, leveraging RGB-D input. To support further research, we also introduced a novel dataset for dynamic scene reconstruction with reliable ground truth. Experimental results demonstrate that our method outperforms traditional non-rigid SLAM approaches.

\paragraph{Limitations:} Our method has primarily been tested on small-scale scenes; extending it to complex real-world scenarios may require 2D priors like point tracking or optical flow. The current implementation runs at 1.5 fps, limiting real-time use. Developing interactive dynamic scene scanning remains important future work.

\section{Acknowledgement}
Research presented in this paper has been supported by
Dyson Technology Ltd. We are very grateful to members of
the Dyson Robotics Lab for their advice and insightful discussions.
\clearpage
\setcounter{page}{1}
\maketitlesupplementary

\setlength{\abovedisplayskip}{1pt} 
\setlength{\belowdisplayskip}{1pt}

We encourage readers to watch the supplementary video for additional details and qualitative results.

\section{Implementation Details}
\subsection{System Details and Hyper parameters}
\paragraph{Non-Rigid SLAM:}
We set the learning weights as follows: $\lambda_{p}=0.9$, $\lambda_{g}=0.1$, $\lambda_{iso}=10.0$ and $\lambda_{n}=0.002$. For the ARAP regularization~\cite{luiten2023dynamic}, we use a nearest neighbor count of 20, a radius of 0.05, and an exponential decay weight of 500. Keyframes are selected with $N=1$. For the MLP, we use an 8-layer architecture with 256 neurons per layer. Frequency encoding is set to 1 for time and 4 for position. MLP is implemented with CUDA-optimized CutlassMLP in tiny-cuda-nn~\cite{tiny-cuda-nn} for the fast optimization.

\paragraph{Static SLAM Ablation:}
We followed the same hyperparameters as MonoGS~\cite{Matsuki:Murai:etal:CVPR2024}, but we use normal loss $L_n$ with the weight $\lambda_n=0.01$ for the entire mapping process and $\lambda_g=0.5$ for the final refinement. For the Replica 3D reconstruction  evaluation, we have used the script introduced in~\cite{Sandström2023ICCV}.

\paragraph{Offline Non-rigid RGB-D Reconstruction Ablation:}
Camera poses are provided by the dataset and remain fixed during training. For the MLP, we adopt the same architecture described in ~\cite{yang2023deformable3dgs}, consisting of an 8-layer network with 256 dimensions per layer, where a concatenated feature vector is input to the fourth layer. The positional encoding frequencies are set to 6 for time and 10 for position. Following the approach in ~\cite{Cai2022NDR, wang2024morpheus}, we evaluate the geometric and appearance metrics against the input views and report the average values.

\section{Camera Pose Jacobian}
We provide the detail of the derivation of camera pose jacobian of 2D Gaussian Splatting in ~\ref{subsec:jacobian}.

We use the notation from~\cite{Sola:etal:ARXIV2018}. Let $\cam \in \SE{3}$ and $\tauC = (\rhoC, \thetaC) \in \se{3}$, the left-side partial derivative on the manifold is defined as:
\begin{equation}
    \mpd{f(\cam)}{\cam} \triangleq  \lim_{\tauC \to 0}\frac{\Log(f(\Exp(\tau) \circ \cam) \circ f(\cam)^{-1})}{\tauC}
\end{equation}

\paragraph{Eq~\ref{eqn:grad_tcw}: }
\begin{align}
\cam &= \Exp(\tauC) = \exp(\tauC^\wedge) \notag \\
     &= \exp\left(\sum_{j=1}^{6} \textbf{E}_j \tau_j\right), \quad j = 1, \dots, 6, \quad \tauC \in \mathbb{R}^6.
\end{align}

where the matrices $\textbf{E}_j \in \mathbb{R}^{4\times 4}$ are the $\SE{3}$ \emph{group generators} and form a basis for $\se{3}$:
\begin{equation}
\begin{aligned}
\textbf{E}_1 &= 
\begin{bmatrix}
       0 & 0 & 0 & 1\\
       0 & 0 & 0 & 0\\
       0 & 0 & 0 & 0\\
       0 & 0 & 0 & 0
\end{bmatrix} \quad
\textbf{E}_2 = 
\begin{bmatrix}
       0 & 0 & 0 & 0\\
       0 & 0 & 0 & 1\\
       0 & 0 & 0 & 0\\
       0 & 0 & 0 & 0
\end{bmatrix} \quad \\ 
\textbf{E}_3 &= 
\begin{bmatrix}
       0 & 0 & 0 & 0\\
       0 & 0 & 0 & 0\\
       0 & 0 & 0 & 1\\
       0 & 0 & 0 & 0
\end{bmatrix} \quad
\textbf{E}_4 = 
\begin{bmatrix}
       0 & 0 & 0 & 0\\
       0 & 0 & -1 & 0\\
       0 & 1 & 0 & 0\\
       0 & 0 & 0 & 0
\end{bmatrix} \quad \\ 
\textbf{E}_5 &= 
\begin{bmatrix}
       0 & 0 & 1 & 0\\
       0 & 0 & 0 & 0\\
       -1 & 0 & 0 & 0\\
       0 & 0 & 0 & 0
\end{bmatrix} \quad
\textbf{E}_6 = 
\begin{bmatrix}   
       0 & -1 & 0 & 0\\
       1 & 0 & 0 & 0\\
       0 & 0 & 0 & 0\\
       0 & 0 & 0 & 0
\end{bmatrix}.
\end{aligned}
\label{eq:group_generators}
\end{equation}

We get the partial derivative as follows:
\begin{equation}
 \quad \pd{}{\tau_j}\exp(\tauC^\wedge)\bigg|_{{\tauC}=0} =  \textbf{E}_j, \quad j = 1, \dots, 6.    
\end{equation}

Therefore, the full derivative is given as:
\begin{align}
    \pd{\cam}{\tauC}\bigg|_{{\tauC}=0} 
    &= \cam \pd{\left(\sum_{j=1}^{6} \textbf{E}_j \tau_j\right)}{\tauC}\bigg|_{{\tauC}=0} \notag\\
\end{align}

Since the meaningful elements of the camera $\cam$ is $12$ number variables, we stack the elements for $12\times6$ matrix and we obtain  

\begin{align}
    \pd{\cam}{\tauC}\bigg|_{{\tauC}=0} 
    &= 
    \begin{bmatrix}
    \mathbf{0} & -\bR_{:, 1}^\times \\
    \mathbf{0} & -\bR_{:, 2}^\times \\
    \mathbf{0} & -\bR_{:, 3}^\times \\ 
    \identity & -{\bt}^\times
    \end{bmatrix}.
\end{align}

where ${\bR} \in \SO{3}$ and ${\bt} \in \mathbb{R}^{3}$ denote the rotation and translation parts of $\cam$.

\paragraph{Eq~\ref{eqn:grad_normal}: }

\begin{align}
{\pd{\bn_c}{\tauC}}\bigg|_{{\tauC}=0} = \mpd{\bn_c}{\camCW} 
    &= \lim_{\tauC \to 0}\frac{\Exp(\tauC) \bn_c - \bn_c}{\tauC} \\
    &= \lim_{\tauC \to 0}\frac{(\identity + \tauC^\wedge) \cdot  \bn_c - \bn_c}{\tauC} \\
    &= \lim_{\tauC \to 0}\frac{\tauC^\wedge \cdot  \bn_c}{\tauC} \\
    &= \lim_{\tauC \to 0}\frac{\thetaC^{\times} \bn_c + \rhoC}{\tauC}  \\
    &= \lim_{\tauC \to 0}\frac{-\bn_c^{\times} \thetaC + \rhoC}{\tauC} \\
    &= \begin{bmatrix} \identity & - \bn_c^\times\end{bmatrix}
\end{align}

\section{Sim4D Training/Test Views}

We define the training and test views on a sphere, with its center representing the target object.
In spherical coordinates ($r$, $\theta$, $\phi$), we set $r = 2.0$. The training view is sampled from two arcs on the sphere's surface, defined by $\theta \in [-10^\circ, 10^\circ]$ and $\phi \in [-10^\circ, 10^\circ]$. The test views are sampled from a circle on the sphere's surface that pass through four key points: $(\theta, \phi) = (5^\circ, 0^\circ)$, $(0^\circ, 5^\circ)$, $(-5^\circ, 0^\circ)$, and $(0^\circ, -5^\circ)$. These points are chosen to ensure uniform sampling around the target object while maintaining a clear separation between the training and test views.

\begin{figure}[h!]
  \center
  \includegraphics[width=\linewidth]{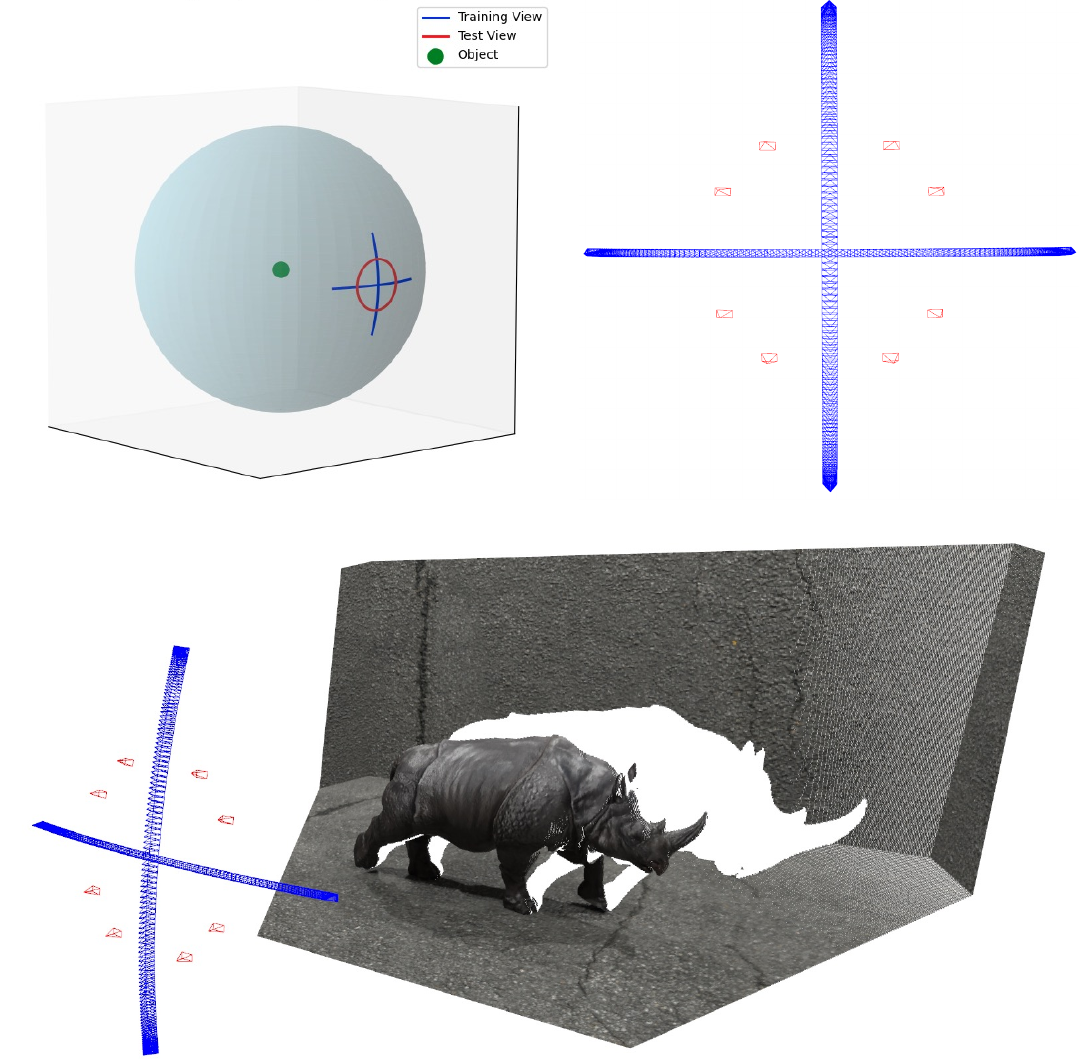}
  \caption{
   \textbf{Training and Test Views on the Sim4D Dataset:} \textbf{Blue} indicates training views, and \textbf{Red} indicates test views. Views are sampled (top right) from an arc on an object-centered sphere (top left) for dynamic scene reconstruction (bottom).}
   \label{fig:view_allocation}
\end{figure}

\section{Further Ablation Analysis}
\subsection{Normal Rigidity Loss}
Table~\ref{table:ablation_normal_rigidity} presents the quantitative results demonstrating the effect of the normal rigidity loss defined in Equation~\ref{eq:normal_rigidity}. The normal rigidity loss improves the overall geometric metrics, such as camera ATE and L1 Depth, for the benchmark sequences by preserving the local geometric consistency of 2D Gaussians.

\begin{table}[h]
\centering
\resizebox{\columnwidth}{!}{%
\begin{tabular}{l|ccccc}
\hline
       & {ATE RMSE} & {L1 Depth} & {PSNR} & {SSIM} & {LPIPS} \\ \hline
\textbf{Ours full}         & \bf{0.28}               & \bf{1.71}                & 28.47                & 0.820            & \bf{0.12}        \\
\textbf{w/o $L_{ARAP\_n}$} & 0.52                   & 2.00                     & \bf{29.04}           & \bf{0.853}      & 0.13             \\ \hline
\end{tabular}
}
\caption{\textbf{Ablation Study on  $L_{ARAP\_n}$.} We report the average number of Sim4D dataset.}
\label{table:ablation_normal_rigidity}
\end{table}

\subsection{Monocular Depth Prior}
While our method was primarily tested with RGB-D camera input, we conducted an ablation study using depth input from the state-of-the-art monocular prediction network~\cite{wang2024moge}, as shown in Table~\ref{table:monocular}. The results demonstrate performance competitive with SurfelWarp, highlighting the potential for purely monocular non-rigid SLAM.

\subsection{Static SLAM Ablation Analysis}
\paragraph{Replica:} Table~\ref{table:rendering_error_replica} shows the photometric rendering performance analysis on the Replica dataset. The results demonstrate that the 2DGS-based SLAM approach offers an advantage in achieving accurate appearance reconstruction.

\paragraph{TUM:} Table~\ref{table:static_tum_combined} presents the full ablation analysis on the TUM dataset. The 2DGS-based approach maintains competitive ATE and appearance metrics while achieving significantly better geometric rendering accuracy, as reflected in the Depth L1 error.

\begin{table}[h]
\centering
\resizebox{0.9\columnwidth}{!}{%
\begin{tabular}{l|l|ccc}
\hline
\textbf{Method}      & \textbf{Metric}        & \textbf{fr1} & \textbf{fr2} & \textbf{fr3} \\ \hline
\multirow{2}{*}{MonoGS} 
                     & ATE RMSE [cm] ↓       & \bf{1.50}    & 1.44         & \bf{1.49}        \\ 
                     & Depth L1 [cm] ↓       & 6.2          & 13.0         & 13.0                  \\ \cline{2-5}
                     & PSNR [dB] ↑           & 23.5         & \bf24.65        & \bf25.09               \\
                     & SSIM ↑                & 0.775        & 0.785         & \bf0.842               \\
                     & LPIPS ↓               & 0.26 1        & \bf0.201        & \bf0.200        \\ \hline
\multirow{2}{*}{MonoGS-2D} 
                     & ATE RMSE [cm] ↓       & 1.58         & \bf{1.2}     & 1.83              \\ 
                     & Depth L1 [cm] ↓       & \bf{3.0}     & \bf{2.3}     & \bf{4.3}         \\ \cline{2-5}
                     & PSNR [dB] ↑           & \bf{23.63}         & 24.47        & 24.05              \\
                     & SSIM ↑                & \bf{0.782}         & \bf0.79         & 0.826            \\
                     & LPIPS ↓               & \bf{0.251}         & 0.228        & 0.223            \\ \hline
\end{tabular}
}
\caption{\textbf{Static SLAM Ablation on TUM Dataset.} Comparison of ATE RMSE, Depth L1, and Rendering Performance Metrics.}
\label{table:static_tum_combined}
\end{table}

\paragraph{Memory Analysis}
Table~\ref{table:memory} presents the average memory usage on the TUM dataset sequences. Due to the geometrically accurate alignment, 2D Gaussians require fewer primitives to represent the scene, resulting in reduced memory consumption.

\begin{table}[h]
\centering
{
\begin{tabular}{cc}
\hline
\multicolumn{2}{c}{\bf{Memory Usage [MB]}} \\
\hline
 \bf{MonoGS-2D}  & MonoGS \\

 \bf{2.73MB}  & 3.97MB \\ 
 \hline 
\end{tabular}
}

\caption{\textbf{Memory Analysis on TUM RGB-D dataset.} 
}
\label{table:memory}
\end{table}

\subsection{Offline Non-Rigid RGB-D Reconstruction Ablation}

Table~\ref{table:comparison_evaluation} provides the full evaluation details of the offline non-rigid RGB-D reconstruction ablation analysis.

\begin{table*}[h]
\centering
\begin{tabular}{lccccccccc}
\hline
 & Metric        & room0      & room1      & room2      & office0      & office1      & office2      & office3      & avg       \\ \hline
\multirow{3}{*}{MonoGS} & PSNR [dB] ↑  & 34.83 & 36.43 & 37.49 & 39.95 & 42.09 & 36.24 & 36.70  & 37.50     \\
                        & SSIM ↑      & 0.954  & 0.959  & 0.9665 & 0.971  & 0.977  & 0.964  & 0.963  & 0.96     \\
                        & LPIPS ↓     & 0.068  & 0.076  & 0.075  & 0.072  & 0.055  & 0.078  & 0.065  & 0.07     \\
 \hline
\multirow{3}{*}{\bf{MonoGS-2D}}   & PSNR [dB] ↑  & \bf36.21 & \bf37.81 & \bf38.7  & \bf43.45 & \bf43.8  & \bf37.48 & \bf37.43 & \bf39.14 \\
                        & SSIM ↑      & \bf0.966  & \bf0.969  & \bf0.9737 & \bf0.985  & \bf0.984  & \bf0.972  & \bf0.971  & \bf0.975 \\
                        & LPIPS ↓     & \bf0.04   & \bf0.042  & \bf0.044 & \bf0.025  & \bf0.029  & \bf0.04   & \bf0.039  & \bf0.038  \\
\hline
\end{tabular}
\caption{Static SLAM Ablation: Rendering Performance Metrics~\cite{Sandström2023ICCV} on Replica Dataset }
\label{table:rendering_error_replica}
\end{table*}

\begin{table*}[h]
\centering
\resizebox{\textwidth}{!}{%
\begin{tabular}{c|>{\centering\arraybackslash}m{2.0cm}|l|cccccccc}
\hline
Method      & Category        & Metric            & curtain & flag & mercedes & modular\_vehicle & rhino & shoe\_rack & water\_effect & wave\_toy \\ \hline
\multirow{5}{*}{Ours (Monocular)}       
                            & \multirow{1}{*}{\centering Trajectory} & ATE RMSE[cm]↓ & {6.23} & {16.29} & {4.90} & {1.86} & {3.17} & {8.02} & {5.52} & {7.21} \\ \cline{2-11} 
                            & \multirow{1}{*}{\centering Geometry} & L1 Depth[cm]↓& 74.2 & 155 & 59.2 & 38.0 & 37.7 & 89.8 & 72.4 & 80.8 \\
                            \cline{2-11}
                            & \multirow{3}{*}{\centering Appearance} & PSNR [dB] ↑       
                            & 17.73 & 16.22 & 20.72 & 26.28 & 21.48 & 17.49 & 18.86 & 17.98 \\
                            &                            & SSIM ↑            & 0.461 & 0.455 & 0.636 & 0.578 & 0253 & 0.448 & 0.390 & 0.441 \\
                            &                            & LPIPS ↓           & 0.297 & 0.517 & 0.282 & 0.380 & 0.339 & 0.391 & 0.258 & 0.281\\ \hline
\end{tabular}
}

\caption{\textbf{Non-rigid SLAM Evaluation on Sim4D Dataset with Monocular Depth Prior.}}
\label{table:monocular}
\end{table*}

\begin{table*}[h]
\centering
\resizebox{\textwidth}{!}{%
\begin{tabular}{|c|l|ccc|ccc|ccc|}
\hline
      &             & \multicolumn{3}{c|}{KillingFusion} & \multicolumn{3}{c|}{DeepDeform} & \multicolumn{3}{c|}{iPhone} \\
      
      &             & frog  & duck  & snoopy & seq002 & seq004 & seq028 & teddy  & mochi  & haru   \\ \hline
\multirow{4}{*}{Morpheus~\cite{wang2024morpheus}} 
            & Depth L1 [cm]     & 4.37  & 3.01 & 2.30  & 2.08  & 1.24  & 2.26  & 5.40  & 0.31  & 1.63  \\ \cline{2-11}
            & PSNR [dB] ↑       & 27.2  & 28.17 & 25.73   & 27.21   & 26.94  & 26.30   & 23.40   & 28.12   & 24.34  \\ 
            & SSIM ↑            & 0.802   & 0.716 & 0.779  & 0.809  & 0.823  & 0.795  & 0.237  & 0.623   & 0.510   \\
            & LPIPS ↓           & 0.31  & 0.419 & 0.483  & 0.301  & 0.428  & 0.397  & 0.776  & 0.55   & 0.564  \\ \hline
\multirow{4}{*}{Ours}     
            & Depth L1 [cm]     & 0.65 & 1.91 & 12.1  & 0.78  & 1.07   & 1.30  & 0.32  & 0.22  & 0.12  \\ \cline{2-11}
            & PSNR [dB] ↑       & 33.72  & 32.75 & 26.95  & 24.36   & 24.13  & 24.02  & 23.89  & 36.15  & 22.60  \\
            & SSIM ↑            & 0.941  & 0.949 & 0.899  & 0.897  & 0.897 & 0.902  & 0.739  & 0.926  & 0.690   \\
            & LPIPS ↓           & 0.063 & 0.073 & 0.257  & 0.245  & 0.313  & 0.241  & 0.259  & 0.131  & 0.391  \\ \hline
\end{tabular}%
}
\caption{\textbf{Offline RGB-D Reconstruction Results}}
\label{table:comparison_evaluation}
\end{table*}

\clearpage
\clearpage

{
    \small
    \bibliographystyle{ieeenat_fullname}
    \bibliography{main, robotvision}
}


\end{document}